\documentclass[letterpaper, 10 pt, conference]{ieeeconf}  

\IEEEoverridecommandlockouts                              

\usepackage{cite}
\usepackage{comment}
\usepackage{amsmath,amssymb,amsfonts}
\usepackage{algorithm}
\usepackage{algpseudocode}
\usepackage{graphicx}
\usepackage{textcomp}
\usepackage{stfloats}
\usepackage{xcolor}
\def\BibTeX{{\rm B\kern-.05em{\sc i\kern-.025em b}\kern-.08em
    T\kern-.1667em\lower.7ex\hbox{E}\kern-.125emX}}
\begin{document}
\title{A Kubernetes-Based Edge Architecture for Controlling the Trajectory of a Resource-Constrained Aerial Robot by Enabling Model Predictive Control}
\author{Achilleas Santi Seisa, Sumeet Gajanan Satpute and George Nikolakopoulos%
\thanks{This work has been partially funded by the European Union's Horizon 2020 Research and Innovation Programme AERO-TRAIN under the Grant Agreement No. 953454.}
\thanks{The authors are with the Robotics and AI Team, Department of Computer, Electrical and Space Engineering, Lule\aa\,\, University of Technology, Lule\aa\,\,}
\thanks{Corresponding Author's email: {\tt\small achsei@ltu.se}}
}
\maketitle
\begin{abstract}
In recent years, cloud and edge architectures have gained tremendous focus for offloading computationally heavy applications. From machine learning and Internet of Thing (IOT) to industrial procedures and robotics, cloud computing have been used extensively for data processing and storage purposes, thanks to its “infinite” resources. On the other hand, cloud computing is characterized by long time delays due to the long distance between the cloud servers and the machine requesting the resources. In contrast, edge computing provides almost real-time services since edge servers are located significantly closer to the source of data. This capability sets edge computing as an ideal option for real-time applications, like high level control, for resource-constrained platforms. In order to utilize the edge resources, several technologies, with basic ones as containers and orchestrators like Kubernetes, have been developed to provide an environment with many features, based on each application's requirements. In this context, this works presents the implementation and evaluation of a novel edge architecture based on Kubernetes orchestration for controlling the trajectory of a resource-constrained Unmanned Aerial Vehicle (UAV) by enabling Model Predictive Control (MPC).

\end{abstract}

\section{Introduction}
\label{intro}
Nowadays, as technology is progressing and the need for computational resources is continuously increasing, different computation layers have been evolved. We can differ these layers into four distinct categories, cloud, fog, edge and devices, while each one of them has its own different characteristics and utilization. At the same time, all of them can be combined with each other to create an ecosystem for the utilization of external computational resources. As these technologies mature, researchers and engineers use them more and more to offload their applications, due to the capabilities and features they provide~\cite{1490177}. Additionally, since the mentioned computation layers have attracted tremendous focus, several state-of-the-art technologies have been developed and are promising to revolutionize many technological fields, like containerized applications and containers' orchestrators. In this framework, robotics can take a huge advantage of the external resources and many resource-constrained platforms can make the most out of them, since they will be able to run algorithms that they can not run on their onboard processors. In this context, edge is emerging since it can provide tremendous resources for enhancing the performance, and the overall efficiency of autonomous operations, and at the same time minimize the travel time delays when transmitting data from UAVs, like in~\cite{100} and~\cite{101}. Thus, edge can be established as a promising solution for time critical operations, like offloading computationally costly controllers, of resource-constrained platforms. In this article, we propose an architecture where we offload the Model Predictive Control method, which is a relatively heavy controller, to the edge, as a containerized application, and we use Kubernetes for managing the containers.

Researchers seek to utilize the advantages of edge computing for the benefit of robots. However, researchers have to overcome some limitations and challenges, in order to use these technologies universally. In~\cite{8934466} an architecture consisting of all four computation layers is used to offload the localization and mapping problem from robots. in this case, edge is operating as a layer between sensor devices, gateways, and cloud servers for enhancing the quality of services, while in~\cite{9090991} edge is used to design a search planner algorithm using deep learning for UAVs. In~\cite{1903.09589} edge and cloud were utilized in terms of storage and computational resources for deep robot learning including object recognition, grasp planning and localization of the computational.

Some works that utilized edge for robotic applications by implementing Kubernetes or container based architectures can be summarized as it follows. In~\cite{kochovski2019architecture} researcher tried to automate, by using Kubernetes orchestration, the process of making decision in terms of placement of the expected workload to edge, fog and cloud, for robotic applications. In~\cite{lumpp2021container}, a methodology based on docker and Kubernetes for ROS-based robotic application is presented. In that case, the architecture was evaluated by experimental results obtained by a mobile robot interacting with an industrial agile production chain.

In the works mentioned above, the approach regarding edge computing is mainly towards non-time critical tasks. High level controllers must operate almost real-time. In~\cite{8812219} an architecture was proposed where the control method consists of the combination of a LQR running on the device and an MPC running the optimization both on edge and cloud, while in~\cite{9304195} two complimentary MPCs are running, one on a local edge and one on the cloud. In comparison to our proposed work, these articles are partly offloading the MPC method on the edge and are focused on evaluating the system in terms of related latency, and the related uncertainty for several cases.

The motivation behind this work is to fill the gap regarding edge enabled robotics. Even though edge computing has proven to be a promising technology to expand the autonomous capabilities of resource-constrained robotic platforms, especially when combined with 5G networks, the research that has been done around this area is relatively limited. Despite the fact that the great advantage of edge computing is the ability of enabling almost real-time operation by offloading the computing process on the edge, most researchers have focused on utilizing edge for offline procedures. Thus, the contribution of this article is to present a novel edge architecture for enabling the time sensitive operation of controlling the trajectory of a resource-constrained UAV in real-time through MPC. Control is one of the basic components of autonomy, thus the performance and efficiency is the main criteria when choosing a controller. Model predictive controllers are widely used on UAVs due to their characteristics and optimal behavior, but they are computationally costly, thus some UAVs, deployed with light processors, like Raspberry Pi, can not handle them. By utilizing the proposed architecture, we will be able to use edge resources in order to offload the MPC and control resource-constrained platforms by closing the loop over the edge. Additionally, we are using Kubernetes orchestration that provides best practices for cloud and edge applications but inserts some challenges that we have to overcome. 

The rest of the article unfolds in the following Sections. In Section~\ref{architecture}, we describe the Kubernetes-based edge architecture, while in Section~\ref{mpc}, we give a brief overview of the UAV and MPC model. In Section~\ref{simulation}, we present the simulation results of the edge architecture in terms of time delays and performance. Finally, in Section~\ref{conclusions}, we conclude the article by highlighting the main points of this work, and we propose future directions.

\section{Kubernetes Edge Architecture}
\label{architecture}
The proposed architecture is based on Kubernetes. Kubernetes is a container orchestrator, developed by Google. Before we start analyzing the Kubernetes-based architecture, we have to describe the containers developed for this work. Afterward, we are going to present the system's architecture and the Robotic Operating System (ROS) framework that was utilized for the UAV-MPC system. Finally, we will describe the communication layer and network.

Containers are based on software that creates an operating environment and are deployed only with the necessary and chosen packages, libraries and dependencies, in order to run a specific application. The application running in this form is called a containerized application. Containers are based on images that are the nominal state of containers before they get deployed. An image can be used to deploy many containers. For our system, we deployed two docker containers. One container is responsible for running the controller and all the necessary libraries and dependencies for its smooth and reliable operation, and the other is responsible for running the ROS master, which takes care of the communication between the ROS nodes. To deploy the two docker containers, we had to developed two different docker images. For both images, we used ROS Noetic on Ubuntu 20.04 entrypoint, and we built on top of them. For the first image, we included several ROS packages and libraries, as well as an optimization engine for the MPC containerized application, while for the second image we just needed to run the ROS master. For a more complex application, we could split it into more containers, each one of them would be assigned a specific task.

\begin{figure}[ht!]
	\centering
	\includegraphics[width=0.9\columnwidth]{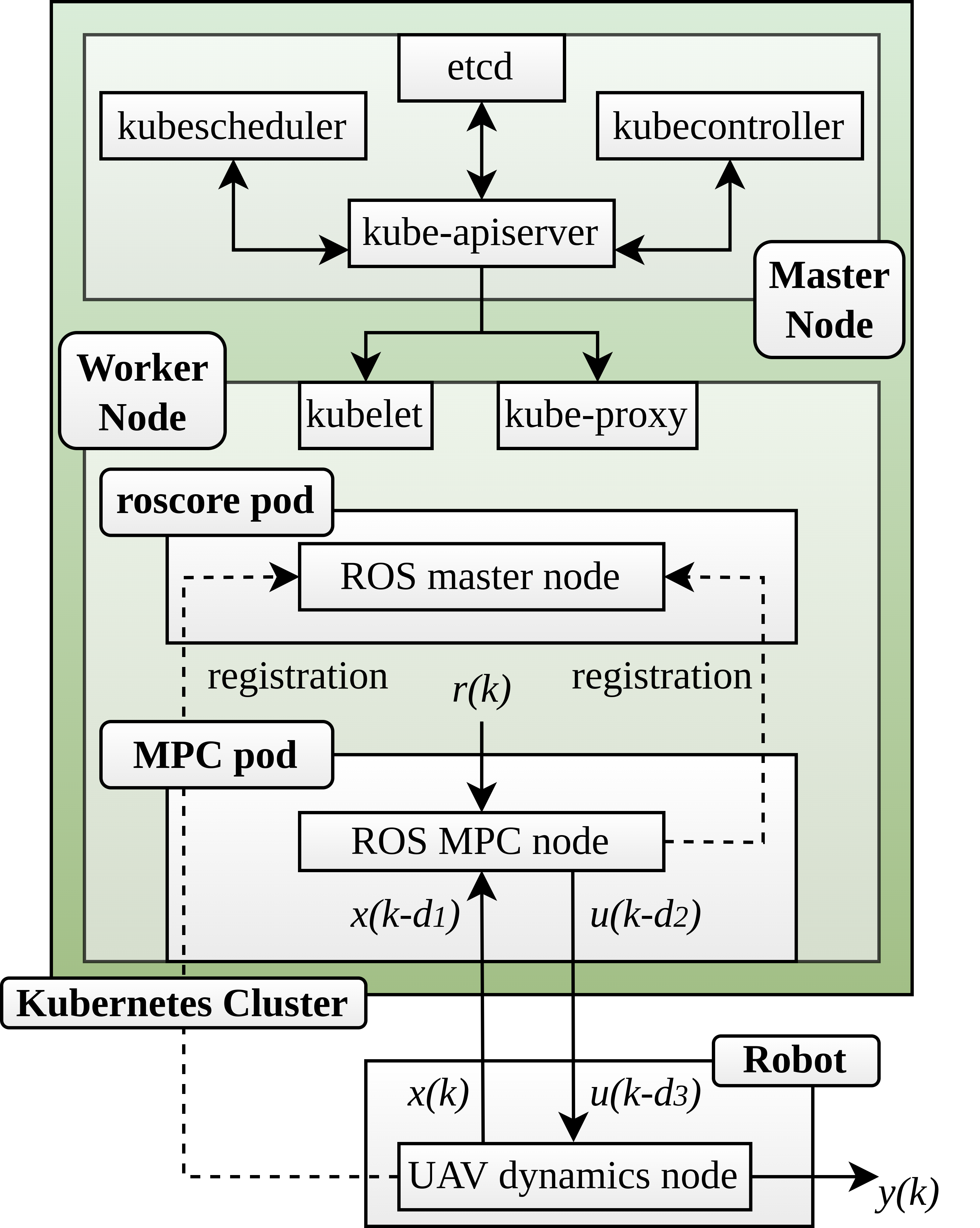}
  	\caption{Diagram of the Kubernetes-based edge architecture for the UAV-MPC system}
  	\label{fig:diagram}
\end{figure}

Once we had developed the docker images, we were able to deploy the docker containers inside the Kubernetes cluster. We decided to use Kubernetes due to the features it provides for our containers. Kubernetes gives us the capability to manage our containers and automates the whole process of deploying the containers, assign them resources, check their health. The services and features that Kubernetes is providing can be extremely helpful for our application, since they give us the chance to manage and monitor our system in an optimal way, and it can even more handful when we have to deploy more containers and the system get more and more complex. The Kubernetes architecture is depicted in Fig.~\ref{fig:diagram}. The top part of the Kubernetes cluster consists of four components that make the master node. These are the kube-apiserver that exposes the Kubernetes Application Programming Interface (API), the etcd that is used as the backing store for all cluster data, the kubescheduler that watches for newly created pods with no assigned node, and selects a node for them to run, and finally the kubecontroller that runs the control processes. Besides the master node, we have the worker nodes. In our case, we have only one worker node, inside which we have deployed our containers in the form of pods. A pod is the basic operational unit of Kubernetes and consist of a set of containers that share storage, network resources, and specifications on how to run the containers. The two pods we have deployed are related to the ROS master and the MPC respectively. Apart from the pods, the worker node consists of the kubelet which makes sure that containers are running in a pod, and kube-proxy which makes sure that network rules are met.

From Fig.~\ref{fig:diagram} we can describe the block diagram of the close loop system. Let's assume that in the time step, $k$ the UAV dynamics node generates a signal $x(k)$ that describes the states of the UAV. These states are the position, velocity, and orientation of the UAV. This signal will arrive at the MPC ROS node, running on the edge, delayed, due to the travel time the signal needs to travel from the UAV to the edge. Thus, the signal carrying the information of $x(k)$ will arrive at the MPC ROS node as $x(k-d_{1})$, while at the same time, another signal regarding the desired states for the UAV will arrive at the MPC ROS node as a reference signal $r(k)$. The controller will have to process this information and generate the command signal $u(k-d_{2})$. Given that $u(k-d_{2})$ is corresponding to the signals $x(k-d_{1})$ and $r(k)$, the variable $d_{2}$ is related to $d_{1}$, as well as the execution time of the MPC. This command signal has to travel from the edge to the UAV in order to close the loop of the system. Thus, the signal arriving to the UAV is denoted as $u(k-d_{3})$, where $d_{3}$ is related to $d_{1}$, $d_{2}$, as well as to the travel time the command signal needs to travel from the edge to the UAV. Finally, the output of the system is denoted as $y(k)$.

The communication between the UAV model simulation and the controller is taken care by ROS. There should be only one ROS master, and every ROS node has to register to that ROS master to be able to run and communicate with other ROS nodes. When two ROS nodes want to exchange data by subscribing and publishing to the same ROS topic, ROS master opens a random port and allows the two ROS nodes to communicate through that port. Once ROS assigns a random ports, different every time, the nodes running on the edge and the nodes running on the robot try to communicate with each other through these ports. Since the containers are deployed on the Kubernetes cluster of the edge machine (host), we have to specify which ports the containers should be exposed to for communication purposes. The challenge occurs because ROS master do not assign specific ports for communication, but it assigns them randomly. To overcome this issue, we used the host network option when we deployed the containers on the Kubernetes cluster, in order to expose all the host ports to the containers and vice versa. That way, the containers can access all the traffic at the host machine's ports and the host machine can access the traffic at the containers' ports. Now, the data coming from the UAV to the edge machine can be forwarded inside the containers and the data from the containerized applications can be exposed to the edge machine and then sent to the UAV.

In this paper, both the edge machine and the UAV are on the same network, thus we were able to use Wi-Fi. Wi-Fi can be an efficient network option for the communication between the UAV and the edge machine and has been used widely, but it is not the optimal solution. 5G is a promising technology that will provide essential features for secure, robust and reliable networking, and can be the field of study for future works.

\section{Model Predictive Control}
\label{mpc}
Model predictive control is a standard method used for high level control for UAVs, thus there are many works describing in detail the behavior of the controller and the kinematics of the UAV, like in~\cite{102}, where authors suggested a UAV model that could afford disturbances by stabilizing its location in space. The preference on MPC in comparison to other common controllers, like PID or LQR, is explained by its predictive behavior and performance. Based on these characteristics, we were prompted to use this controller for controlling the trajectory of an UAV, and we were motivated to offload it to the edge so resource-constrained UAVs and robots in general, that can not afford to run this controller onboard, would be able to take advantage of the benefits of MPC. The UAV model and the implementation of the MPC for this work are based on~\cite{9143931}.

\subsection{UAV Model}
\label{kinematics}
In order to develop the MPC methodology, the first step is to describe the UAV kinematics model, which is presented through the Eq.~\ref{eq:kinematics}.

\begin{align}
&\dot{p}(t) = v_{z}(t) \nonumber\\
&\dot{v}(t) = R_{x,y}(\theta,\phi) \begin{bmatrix} 0\\ 0\\ T\end{bmatrix} + \begin{bmatrix} 0\\ 0\\ -g\end{bmatrix} - \begin{bmatrix} A_{x} & 0 & 0\\ 0 & A_{y} & 0\\ 0 & 0 & A_{z}\end{bmatrix}u(t) \label{eq:kinematics}\\
&\dot{\phi}(t) = \frac{1}{\tau_{\phi}} (K_{\phi} \phi_{d}(t) - \phi(t)) \nonumber\\
&\dot{\theta}(t) = \frac{1}{\tau_{\theta}} (K_{\theta} \theta_{d}(t) - \theta(t)), \nonumber
\end{align} 

\begin{figure}[ht!]
	\centering
	\includegraphics[width=0.9\columnwidth]{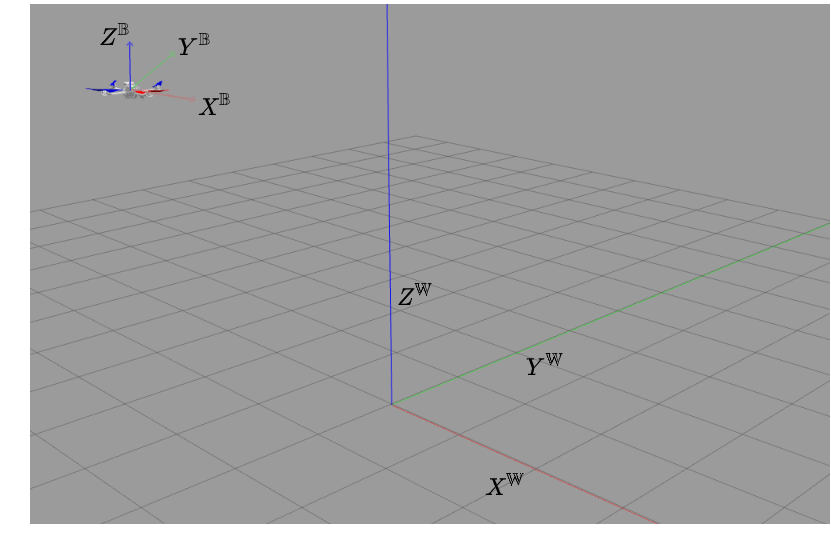}
  	\caption{Coordinate frames, where $\mathbb{W}$ and $\mathbb{B}$ represent the world and body coordinate frames respectively on gazebo simulation environment}
  	\label{fig:simulation}
\end{figure}

where $p = [p_{x}, p_{y}, p_{z}]^{T}$ and $v = [v_{x}, v_{y}, v_{z}]^{T}$ are the position and the linear velocity respectively based on the world frame ($\mathbb{W}$), as depicted in Fig.~\ref{fig:simulation}. We donate as $R(\phi(t), \theta(t)) \in SO(3)$ the rotation matrix that represents the attitude. $\phi$ and $\theta \in [-\pi, \pi]$ are the roll and pitch angles, while $T \geq 0$ describes the total thrust. The acceleration depends on the magnitude and angle of the thrust vector, the gravity, and the linear damping terms $A_{x}, A_{y}, A_{z} \in R$ $g$. $\phi_{d}$ and $\theta_{d} \in R$ are the desired roll and pitch inputs with gains $K_{\phi}$ and $K_{\theta} \in R$, and time constants $\tau_{\phi}$ and $\tau_{\theta} \in R$.

\subsection{Cost Function}
\label{cost_function}
Next step for the MPC methodology is to present the cost function. $x = [p, v, \phi, \theta]^{T}$ and $u = [T, \phi_{d}, \theta_{d}]^{T}$ represent the UAV's state vector and the control input, respectively. The sampling time of the system is $\delta_{t} \in \mathbb{Z}^{+}$, while the forward Euler method is used for each time instance $(k+1|k)$. The predictive behavior of the MPC is based on the prediction horizon, which considers a specified number of steps into the future, and is represented as $N$.

In order to minimize the cost of the cost function, an optimizer has been assigned to find the optimal set of control actions. The cost function associates the cost of the configuration of states and inputs at the current time and in the prediction. $x_{k+j|k}$ represents the predicted states at the time step $k+j$, produced at the time step $k$, while $u_{k+j|k}$ represents the corresponding control actions. Furthermore, $x_{k}$ represents the predicted states and $u_{k}$ represents the corresponding control inputs along the prediction horizon. The equation describing the cost function is presented in Eq.~\ref{eq:cost_funtion}.

\begin{align}
&J = \sum_{j=1}^{N} \underbrace{(x_{d} - x_{k+j|k})^{T} Q_{x} (x_{d} - x_{k+j|k})}_{state \quad cost} \nonumber\\
&+ \underbrace{(u_{d} - u_{k+j|k})^{T} Q_{u} (u_{d} - u_{k+j|k})}_{input \quad cost} \label{eq:cost_funtion}\\
&+ \underbrace{u_{k+j|k} - u_{k+j-1|k})^{T} Q_{\delta u} (u_{k+j|k} - u_{k+j-1|k}),}_{control \quad actions \quad smoothness \quad cost} \nonumber
\end{align} 

\begin{figure*}[b]
	\centering
	\includegraphics[width=\textwidth]{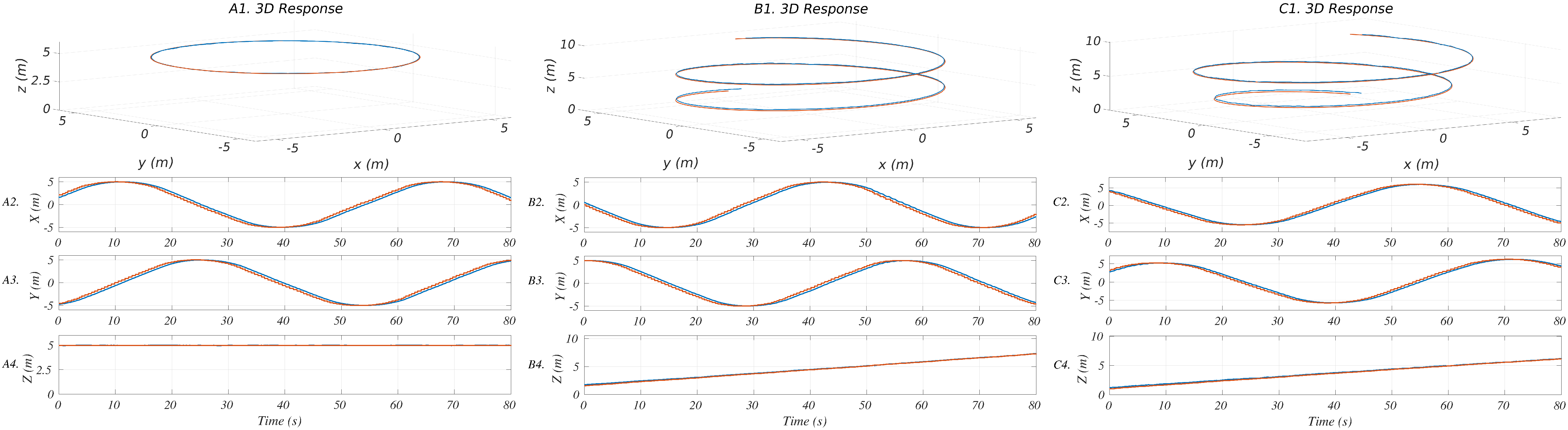}
  	\caption{UAV responses based on the Kubernetes-based architecture for circular, spiral and helical trajectories. A1)Depicts the 3D response of the circular trajectory while A2, A3, A4 depict the responses of the $X$, $Y$ and $Z$ axis respectively. B1) Depicts the 3D response of the spiral trajectory. C1) Depicts the 3D response of the helical trajectory. The blue line represents the real trajectory of the UAV, while the blue line represents the reference points for the desired trajectory of the UAV.}
  	\label{fig:responses}
\end{figure*}

where the first term denotes the cost related to the deviation between the predicted states and the certain desired states $x_{d}$, while $Q_{x} \in \mathbb{R}^{8x8}$ is a matrix describing the state weights. The second term denotes the input cost, describing hovering, and that penalizes a deviation from the steady-state input $u_{d} = [g, 0, 0]$, while $Q_{u}$ is a matrix describing the input weights. The third term is added to guarantee that the control actions are smooth. This is achieved by comparing the input at $(k+j-1|k)$ with the input at $(k+j|k)$ and penalizing the changing of the input from one time step to the next one, with $N \in N^{+}$ to denote the control Horizon of the MPC, while $Q_{\delta u} \in \mathbb{R}^{3x3}$ is a matrix describing the input rate weights.

\section{Simulation Results}
\label{simulation}
In this section, we are presenting the simulation results of the proposed architecture. For the simulation, we used the gazebo environment and the UAV simulation model hummingbird of the rotor simulation ROS package, as depicted in Fig.~\ref{fig:simulation}. For the edge, we utilized a powerful machine and microk8s was running on the edge, which is a lightweight Kubernetes software that was used as the Kubernetes orchestrator. The specifications of the edge are: 1) Processor: Intel Core i5-8400 CPU@2.80GHz×6, 2) Memory: 32 GB 3) Operating System: Ubuntu 20.04 LTS and 4) Disk Capacity: 2.5 TB.

For the following simulations, the MPC horizon was set at 100 steps and the MPC rate was set at 100Hz. We were able to select this values, because the MPC is running on the edge and we are using its capabilities. UAVs' onboard processors would not be able to handle an MPC with these high values since they increase the complexity of the controller (solution of the optimization problem), thus the computational demands.

In Fig.~\ref{fig:responses}, the responses for the three different tested trajectories of the UAV are depicted. The first line of figures depicts the 3D response of the circular, spiral and helical trajectories while the second, third and fourth lines depict the responses of the $X$, $Y$ and $Z$ axis respectively, for each different trajectory. The blue line represents the real trajectory of the UAV, while the blue line represents the reference points for the desired trajectory of the UAV. From these figures, we can notice that the UAV simulation model can successfully follow the desired trajectory. The time delays seem to not have a significant effect on the performance of the controller. On the next figures, we are investigating in more detail these time delays.

\begin{figure}[ht!]
	\centering
	\includegraphics[width=\columnwidth]{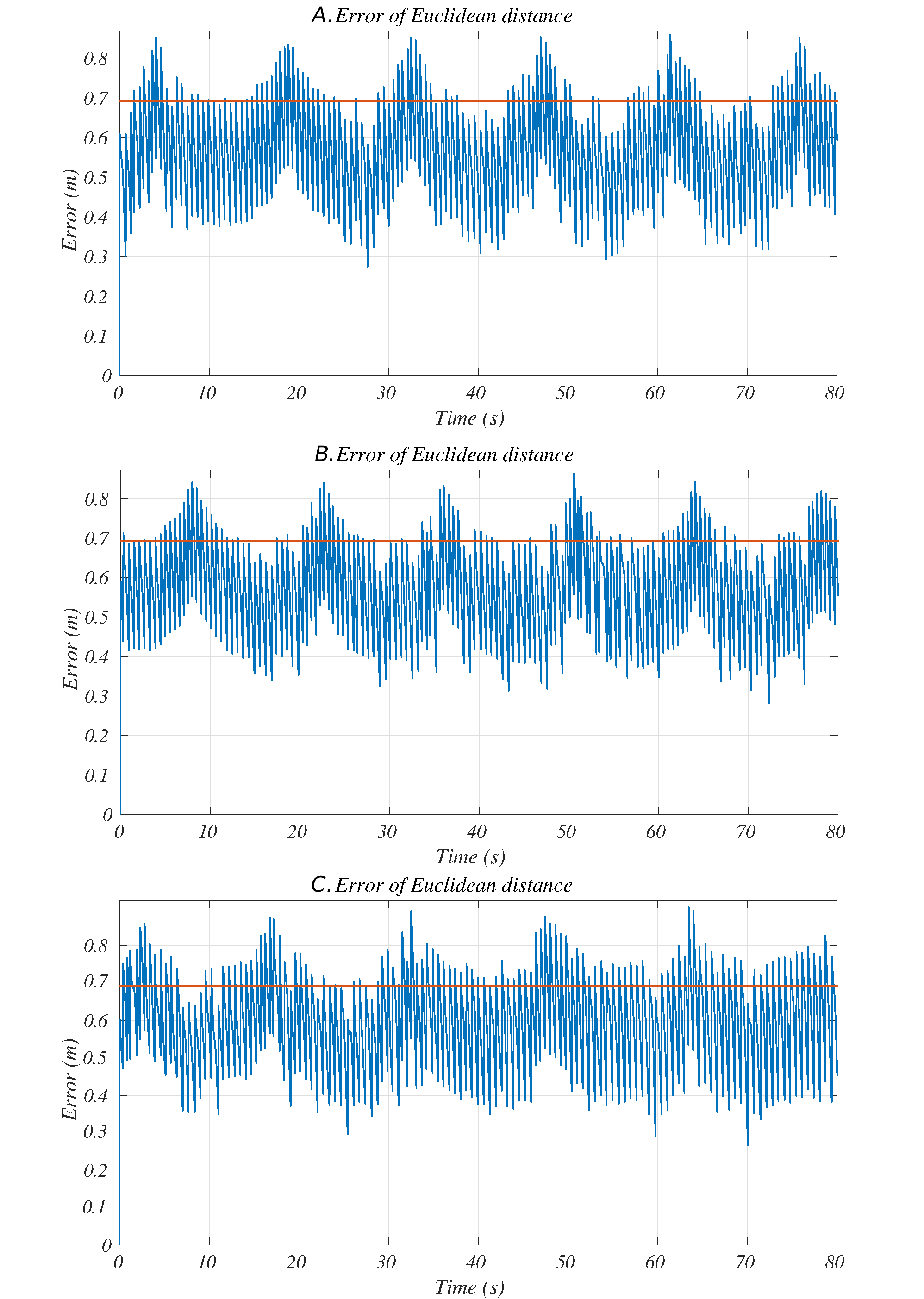}
  	\caption{Euclidean error between UAV position and reference point for each time step of the Kubernetes-based architectures, for A) the circular, B) spiral, and C) helical trajectory. The blue line represents the error and the red line represents the error tolerance}
  	\label{fig:error}
\end{figure}

\begin{figure*}[b]
	\centering
	\includegraphics[width=\textwidth]{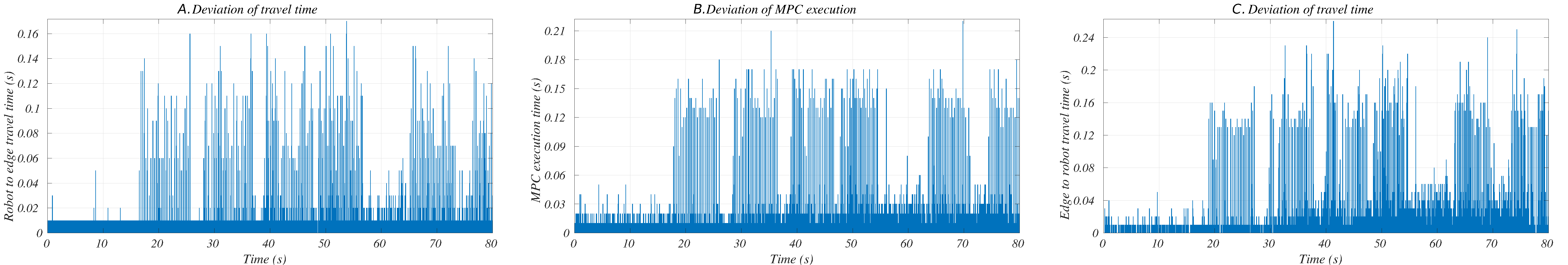}
  	\caption{Deviation of the different types of time delays for the spiral trajectory: A) Deviation for the travel time of a signal from the UAV to the edge. B) Deviation for the execution time of the MPC. C) Deviation for the travel time of a signal from the edge to the UAV.}
  	\label{fig:times}
\end{figure*}

Fig.~\ref{fig:error} depicts the Euclidean error between the UAV position and the reference point for each time step of the Kubernetes-based architectures, for the circular, spiral and helical trajectories. The blue line represents the error and the red line represents the error tolerance. The controller is responsible to keep the error below the tolerance value. If the error goes above the tolerance, the controller will correct it and the UAV will continue following the desired trajectory. The tolerance was set at 0.4 meters for each axis, thus in total of $\sqrt{0.68}$ meters.

In Fig.~\ref{fig:times}, the deviation of the different types of time delays for the spiral trajectory are presented. In the left figure, the deviation for the travel time of a signal from the UAV to the edge, in the middle figure the deviation for the execution time of the MPC, and in the right figure the deviation for the travel time of a signal from the edge to the UAV, are depicted. The average measured travel time from the UAV to the edge is 0.0089 seconds, and the maximum 0.1700 seconds. For the execution time, the average measured time is 0.0141 seconds and the maximum is 0.2200 seconds. Finally, for the travel time, from the edge to the UAV, the measured travel time is 0.0161 seconds and the maximum is 0.2600 seconds.

\begin{figure}[ht!]
	\centering
	\includegraphics[width=\columnwidth]{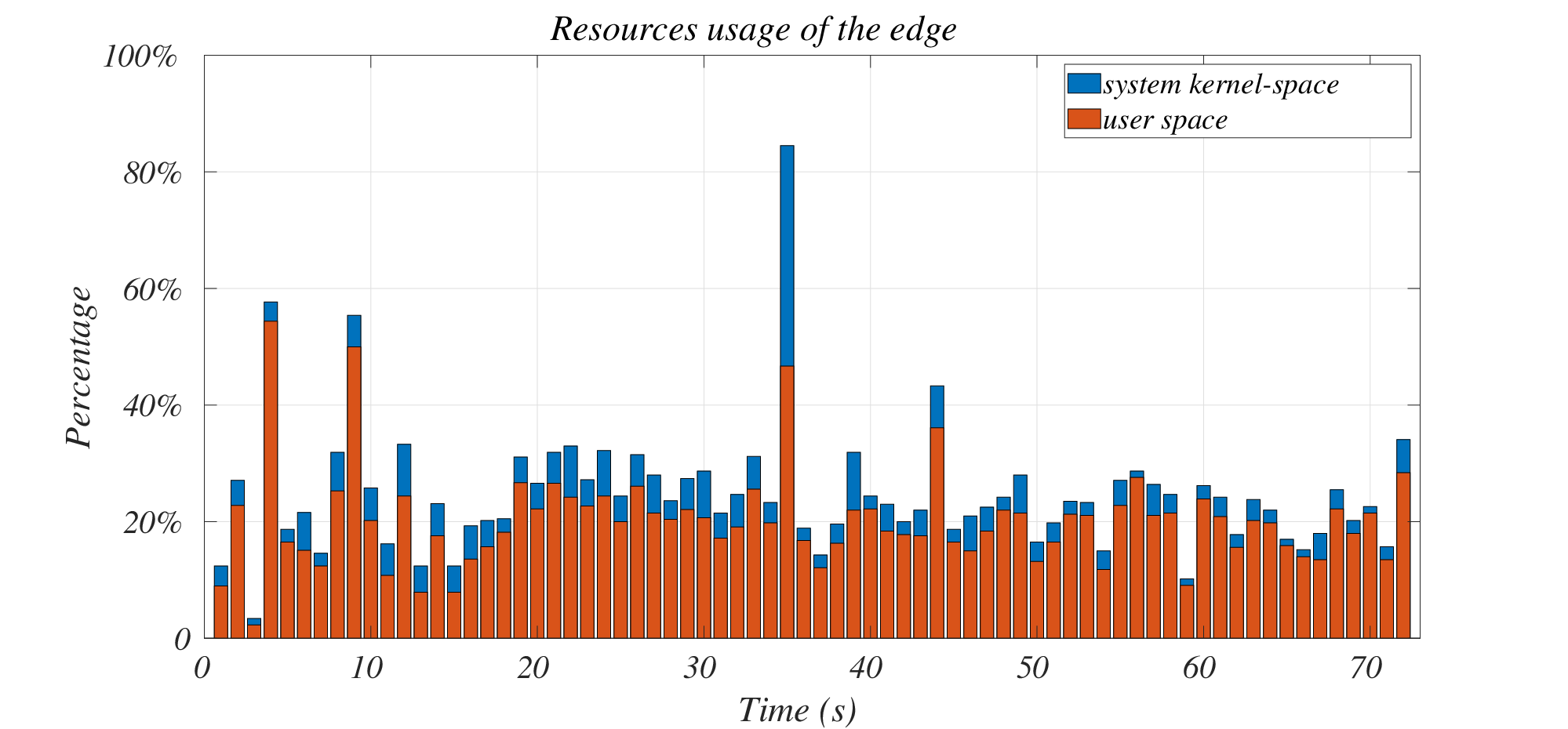}
  	\caption{Edge resources usage during the spiral trajectory. The red bar represents the user space and the blue bar represents the system kernel-space.}
  	\label{fig:resources}
\end{figure}

To end the evaluation of the system, we measured the resource usage for the execution of the MPC on the edge and the data are depicted in Fig.~\ref{fig:resources}. The red bars represent the time the CPU spends executing processes in user-space (us). Similarly, the blue bar represents the time spent on running system kernel-space (sy) processes. From the figure we can observe that by utilizing the edge machine, the edge does not get overloaded, and the maximum reached value is 84.50\% which occurs when the values us and sy are 46.70\% and 37.80\% respectively. The maximum values that us and sy reach independently are 54.40\% and 37.80\% respectively, and their average values are 20.225\% for the us and 4.582 for the sy. From these measurements and figure, we can notice that the relatively immense assigned edge resources are adequate in order to run the computationally demanding controller, but even in this case, during the $35^{th}$ second of the trajectory, the usage of resources were almost at $90\%$. This means that computational light units, like UAVs' onboard processors, might not be able to execute that controller smoothly. 

\section{Conclusions and Future Work}
\label{conclusions}
In this work, we presented a novel edge architecture to control the trajectory of an UAV through the edge by enabling an MPC methodology. This architecture can be beneficial for expanding the computational capabilities of resource-constrained platforms like aerial robots, that in many cases are deployed with light microprocessors onboard, like Raspberry Pi, and can not afford to run computationally expensive processes onboard. By utilizing edge, we were able to offload the controller there, and control the trajectory of the UAV in real-time by closing the loop of the system through the edge. Furthermore, we evaluated the proposed architecture, through a series of experiments, through which we examined the performance of the system, as well as the overall time delays.

Edge computing is a promising technology for the field of robotics. In the current article, we offloaded the computationally costly MPC, while future works can move towards offloading other time sensitive robotic application, like sensor fusion for online perception, or offload applications that require many resources in order to operate in real-time, like map merging from multiple agents. The end goal would be to create an ecosystem through which multiple agents will be able not only to use edge resources to expand their autonomy capacity, but also communicate and collaborate through the edge.

\bibliographystyle{./IEEEtranBST/IEEEtran}
\bibliography{./IEEEtranBST/IEEEabrv,references}

\end{document}